\documentclass{article} 
\usepackage{nips14submit_e,times}
\usepackage{hyperref}
\usepackage{url}
\usepackage{amsfonts,amssymb,amsthm}
\usepackage{subfigure}
\usepackage{graphicx}
\usepackage{mdwlist}
\usepackage{algorithm}
\usepackage{algorithmic}
\usepackage{booktabs}
\usepackage{amsmath}
\usepackage{blkarray}
\usepackage{multirow}
\usepackage{mathtools}
\usepackage{graphicx}
\usepackage{enumerate}
\usepackage{enumitem}
\usepackage{ textcomp }
\usepackage{ wasysym }
\usepackage[latin1]{inputenc}
\usepackage{tikz}
\usetikzlibrary{arrows,positioning,shapes.geometric}

\title{Learning computationally efficient dictionaries and their implementation as fast transforms}

\author{
Luc Le Magoarou \qquad \quad R\'emi Gribonval 
\\
Inria\\
Centre Inria Rennes - Bretagne Atlantique\\
\texttt{\{luc.le-magoarou,remi.gribonval\}@inria.fr} \\
}

\nipsfinalcopy 

\begin{document}
\setlength\abovecaptionskip{0.4ex}
\setlength\belowcaptionskip{0.2ex}

\maketitle

\begin{abstract}
Dictionary learning is a branch of signal processing and machine learning that aims at finding a frame (called dictionary) in which some training data admits a sparse representation. The sparser the representation, the better the dictionary. The resulting dictionary is in general a dense matrix, and its manipulation can be computationally costly both at the learning stage and later in the usage of this dictionary, for tasks such as sparse coding. Dictionary learning is thus limited to relatively small-scale problems. In this paper, inspired by usual fast transforms, we consider a general dictionary structure that allows cheaper manipulation, and propose an algorithm to learn such dictionaries --and their fast implementation-- over training data. The approach is demonstrated experimentally with the factorization of the Hadamard matrix and with synthetic dictionary learning experiments. 
\end{abstract}

\section{Introduction}
Sparse representations using dictionaries are a popular way of providing concise descriptions of high-dimensional vectors.
The goal of dictionary learning is to find an appropriate dictionary $\mathbf{D}$ allowing the sparse approximation of a training collection, gathered in a data matrix $\mathbf{X}$, as:
\begin{equation}
\mathbf{X} \approx \mathbf{D}\boldsymbol{\Gamma},
\end{equation}
where 
$\boldsymbol{\Gamma}$ has sparse columns. 
Historically, the only way to come up with a dictionary was to analyse mathematically the data and derive a "simple"  formula to construct the dictionary. 
Dictionaries designed this way are called \emph{analytic dictionaries}  \cite{Rubinstein2010} (e.g.: associated to Fourier, wavelets and Hadamard transforms). Due to the relative simplicity of analytic dictionaries, they are often associated with a fast algorithm such as the Fast Fourier Transform (FFT) \cite{CooleyTukey1965} or the Discrete Wavelet Transform (DWT) \cite{Mallat1989}.
On the other hand, the development of modern computers allowed the surfacing of automatic methods that learn a dictionary directly from the data. 
Such \emph{learned dictionaries} are usually well adapted to the data at hand, but due to their lack of structure, they do not lead to fast algorithms and are costly to store. A survey on dictionaries, analytic or learned, can be found in \cite{Rubinstein2010}. 

Can one design dictionaries as well adapted to the data as learned dictionaries, while as fast to manipulate and as cheap to store as analytic ones? Such an objective can seem unrealistic, but in \cite{Rubinstein2010a}, and more recently in \cite{Chabiron2013}, the authors introduced new dictionary structures that seek to bridge the gap between the two categories. The model we introduce actually generalizes these approaches. We build on the simple observation that the fast transforms associated with analytic dictionaries can be seen as consecutive multiplications of the input vector by sparse matrices, indicating that such dictionaries can be expressed as a product of sparse matrices\footnote{The product being taken from left to right: $\prod_{i=1}^N \mathbf{A}_i = \mathbf{A}_1 \cdots \mathbf{A}_N$}: 
\begin{equation}
\mathbf{D} = \prod_{j=1}^M \mathbf{S}_j. 
\label{eq:dicfac}
\end{equation}

This factorizable structure is precisely what enables fast algorithms to multiply a vector by the dictionary or its adjoint. For example each step of the butterfly radix-2 FFT can be seen as the multiplication by a sparse matrix having only two non-zero entries per row an per column, which leads to the well-known complexity savings. Another example can be found in Figure~\ref{fig:hadamard_true} where we show the Hadamard dictionary along with its factorized form.

Our objective is thus to learn a factorizable dictionary  (i.e.\ taking the form or eq.\eqref{eq:dicfac}), making it intrinsically fast to manipulate and cheap to store. We will express this as an highly non-convex optimization problem, and rely on recent advances in optimization such as the PALM algorithm proposed in \cite{Bolte2013} to address it. 
In Section~\ref{sec:relatedworks} we formulate the problem 
 and link our work with relevant others, in Section~\ref{sec:optimframework} we present a general algorithm to solve it, and finally in Section~\ref{sec:experiments} we present experimental results showing the interest of the proposed method. In particular, we demonstrate its ability to factor the Hadamard matrix in a way that enables its multiplication by an arbitrary vector as efficiently as with the fast Hadamard transform.

\begin{figure}[tbp]
\includegraphics[width=1\textwidth]{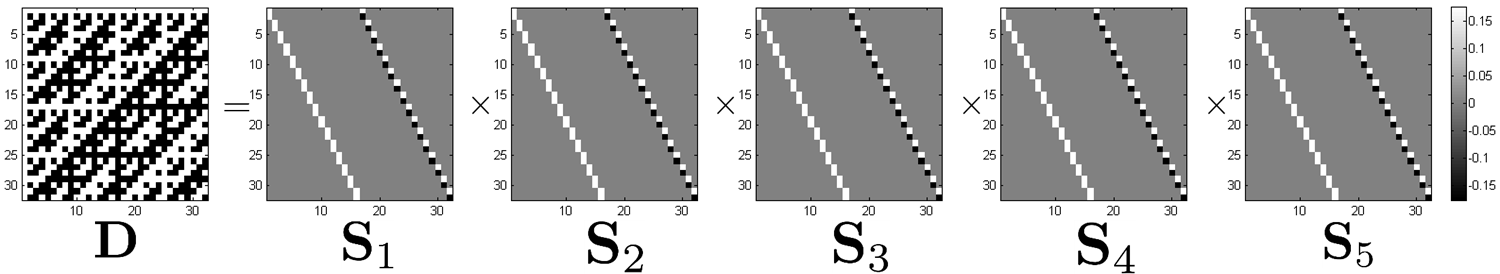}
\caption{The Hadamard dictionary in size $n\times n$ with $n=32$ (left) and its factorization. The dictionary is totally dense so that the naive storage and multiplication cost $\mathcal{O}(n^2=1024)$. On the other hand, we show the factorization of the dictionary in $\log_2(n) = 5$ factors, each having $2n = 64$ non-zero entries, so that the storage and multiplication in the factorized form cost \mbox{$\mathcal{O}(2n\log_2(n)=320)$.}}
\label{fig:hadamard_true}
\end{figure}

\section{Problem formulation and related works}
\label{sec:relatedworks}
\paragraph{Notation.} Throughout this paper, matrices are denoted by bold upper-case letters: $\mathbf{A}$. Vectors are denoted by bold lower-case letters: $\mathbf{a}$. The $i$th column of a matrix $\mathbf{A}$ is denoted by: $\mathbf{a}_i$. 
Sets are denoted by calligraphical symbols: $\mathcal{A}$. The standard vectorization operator is denoted by $\text{vec}(\cdot)$ and the Kronecker product by $\otimes$. The usual $\ell_0$-norm is denoted by $\left\Vert\cdot\right\Vert_0^0$ (it counts the number of non-zero elements), $\left\Vert\cdot\right\Vert_F$ denotes the Frobenius norm, and $\left\Vert\cdot\right\Vert_{2}$ the operator norm.

\paragraph{Objective.} As stated in the introduction, our goal is to learn dictionaries that are intrinsically fast to manipulate and cheap to store. In order to meet these requirements, we impose that our dictionary be the product of several sparse matrices. Mathematically speaking, let $\mathbf{X} \in \mathbb{R}^{d \times n}$ be our data matrix, each of its $n$ columns $\mathbf{x}_i$ being a training vector, $\mathbf{D} \in \mathbb{R}^{d \times a}$ be our dictionary with $a$ atoms and \mbox{$\boldsymbol{\Gamma} \in \mathbb{R}^{a \times n}$} be the corresponding sparse representation matrix such that \mbox{$\mathbf{X} \approx \mathbf{D}\boldsymbol{\Gamma}$}. In order to meet the requirements and be intrinsically fast, $\mathbf{D}$ must take the form of eq.\eqref{eq:dicfac}, where the $\mathbf{S}_j$s are sparse matrices in $\mathbb{R}^{a_{j} \times a_{j+1}}$ with $a_{1} = d$ and $a_{M+1} = a$. Then, denoting   $\mathbf{S}_{M+1} = \boldsymbol{\Gamma}$ for ease of notation, our goal is to find the sparse factors $\mathbf{S}_j$s such that: 
\begin{equation}
\mathbf{X} \approx \prod_{j=1}^{M+1} \mathbf{S}_j.
\label{eq:matfact}
\end{equation}
Under this form, our problem amounts to a factorisation of the data matrix into $M+1$ sparse factors, thus it can be cast as a general optimization problem:
\begin{equation}
\begin{array}{c}
\underset{\mathbf{S}_1, \ldots,\mathbf{S}_{M+1}}{\text{Minimize }}\quad   d\big(\mathbf{X} , \prod\limits_{j=1}^{M+1}\mathbf{S}_j\big) + \sum\limits_{j=1}^{M+1}g_j(\mathbf{S}_j),
\end{array}
\label{eq:verygeneralproblem}
\end{equation}
where $d(\cdot,\cdot)$ is some distance measure and the $g_j$s are sparsity-seeking penalties or constraints. This general optimization problem has been studied recently by various authors in several    domains.

\paragraph{Related works.}For dictionary learning, as mentioned in the introduction, two main works have begun to explore this way. In \cite{Rubinstein2010a}, the authors propose to learn a dictionary which atoms are sparse linear combinations of atoms of a so-called \emph{base dictionary}. The base dictionary should be associated with a fast algorithm (it takes the form of eq.\eqref{eq:dicfac}), so that the whole learned dictionary can be efficiently stored and manipulated. It can be seen as having the $M-1$ leftmost factors fixed in eq.\eqref{eq:matfact} (let us call it $\mathbf{D}_{\textrm{base}}$), the $M$th factor being the sparse representation of the dictionary over the base dictionary ($\mathbf{D} = \mathbf{D}_{\textrm{base}} \mathbf{S}_M$), and the $M+1$th being the sparse representation of the training data over the learned dictionary. The major drawback with this formulation is that the learned dictionary is highly biased toward the base dictionary, so that we do not have full adaptability. In \cite{Chabiron2013}, the authors propose to learn a dictionary in which each atom is the composition of several circular convolutions with sparse kernels, so that the dictionary is fast to manipulate. Their model can be seen as eq.\eqref{eq:matfact}, with the $g_j$s corresponding to the $M$ leftmost factors imposing sparse circulant matrices. This formulation is limited in nature to the case where the dictionary is well approximated by a product of sparse circulant matrices.

In statistics and data analysis, some researchers have been interested in statistical models in which the covariance matrix of the data takes the form of eq.\eqref{eq:matfact}, so that estimating this covariance matrix amounts to the problem of eq.\eqref{eq:verygeneralproblem}. Recent representative works in this direction are \cite{Lee2008} and \cite{Cao2011}.

Even more recently, similar models were proposed in machine learning. In \cite{Lyu2013}, the authors introduce the sparse multi-factor NMF, that can be seen as modelling the data as in eq.\eqref{eq:matfact}, with all $\mathbf{S}_j$s being non-negative matrices. In \cite{Neyshabur2013} and \cite{Arora2013}, the authors assume that the data come from a deep neural network, assuming that consecutive layers are sparsely connected and neglecting the non-linearities, they provide some strategies to recover the structure of the network. This model can be seen as modelling the data like in eq.\eqref{eq:matfact} with the $M$ leftmost factors representing a layer of the network each (the non-linear part being omitted), and the $M+1$th factor being the input of the network.

Note that another concern, somewhat related to that of having a computationally efficient dictionary, is that of being able to rapidly compute the sparse code $\boldsymbol{\Gamma}$ corresponding to the training data $\mathbf{X}$ given the dictionary $\mathbf{D}$. Models addressing this problematic have been proposed in \cite{Gregor2010} and \cite{Sprechmann2012}.

\section{Optimization framework}
\label{sec:optimframework}
In this section we explicit the considered optimization problem, and describe an algorithm that is guaranteed to converge to a stationary point of the objective function.

\subsection{Objective function}
\label{ssec:objective}
To learn a dictionary that is well adapted to the data while being fast to manipulate and cheap to store, we will minimize an objective function of the form of eq.\eqref{eq:verygeneralproblem}. We will take as distance measure the squared Frobenius norm of the difference $ d(\mathbf{X} , \prod_{j=1}^{M+1}\mathbf{S}_j):=\frac{1}{2} \Vert\mathbf{X} - \prod_{j=1}^{M+1}\mathbf{S}_j\Vert_F^2 $ and as sparsity-seeking penalties some indicator functions of sets of sparse matrices: $g_j:=\delta_{\mathcal{E}_j}$, with $\delta_{\mathcal{T}}(\mathbf{S}) = 0$ if $\mathbf{S}\in\mathcal{T}$ and $\delta_{\mathcal{T}}(\mathbf{S}) =\infty$ otherwise.
The keen reader might have noticed that this basic formulation of the objective is invariant under relative scalings of the factors if the constraint sets are scale invariant themselves, and we address this issue below.

\paragraph{Choice of the constraints.}
\label{ssec:constraintchoice}
 The choice of the constraint sets is crucial, because they entirely determine the storage and multiplication cost of the learned dictionary. Indeed, storing/multiplying the dictionary in the factorized form will cost $\mathcal{O}(\sum_{j=1}^{M}\left\Vert \text{vec}(\mathbf{S}_j)\right\Vert_0^0)$, whereas classical dictionary learning methods would typically provide dense dictionaries for which storing/multiplying would cost $\mathcal{O}(da)$. This simple statement allows to introduce the \emph{Relative Complexity} (RC) of the learned dictionary:
\begin{equation}
\text{RC}:= \frac{\sum_{j=1}^{M}\left\Vert \text{vec}(\mathbf{S}_j)\right\Vert_0^0}{da}.
\label{economy}
\end{equation}
This quantity is clearly positive and should be smaller than $1$ in order to make complexity savings. In practice, we will usually choose $\mathcal{E}_i$s that are subsets of "$\ell_0$ balls", namely they will take the form: $\mathcal{E}_j = \mathcal{N}_j\cap\{\mathbf{A} \in \mathbb{R}^{a_j \times a_{j+1}} : \left\Vert \text{vec}(\mathbf{A}) \right\Vert^0_0 \leq p_j\},$ where $\mathcal{N}_j$ is an arbitrary set imposing additional constraints. These constraints will give us: $\text{RC} \leq  \sum_{j=1}^{M}p_j/da$.

\paragraph{Coping with the scaling ambiguity.}
\label{ssec:scalinganbiguity}
In order to avoid scaling ambiguities, it is common \cite{Chabiron2013,Lyu2013} to normalize the factors and introduce a multiplicative scalar $\lambda$ in the data fidelity term. Doing so, the actual problem that we consider is the following:
\begin{equation}
\begin{array}{c}
\underset{\lambda,\mathbf{S}_1, \ldots,\mathbf{S}_{M+1}}{\text{Minimize }}\quad \Psi(\mathbf{S}_1, \ldots ,\mathbf{S}_{M+1},\lambda) := \frac{1}{2} \Big\Vert\mathbf{X} - \lambda\prod\limits_{j=1}^{M+1}\mathbf{S}_j\Big\Vert_F^2 + \sum\limits_{j=1}^{M+1}\delta_{\mathcal{E}_j}(\mathbf{S}_j),
\end{array}
\label{eq:actualproblemscal}
\end{equation}
with a new canonical form for the $\mathcal{E}_j$s namely $\mathcal{E}_j = \mathcal{N}_j\cap\{\mathbf{A} \in \mathbb{R}^{a_j \times a_{j+1}} : \left\Vert \text{vec}(\mathbf{A}) \right\Vert^0_0 \leq p_j, \left\Vert \mathbf{A} \right\Vert_F = 1\}$, so that the factors are normalized.

\subsection{Algorithm overview}
\label{ssec:algorithmoverview}
The formulation of the problem in eq.\eqref{eq:actualproblemscal} is unfortunately highly non-convex, and the sparsity enforcing part is non-smooth. Stemming on recent advances in non-convex optimization, we propose next an algorithm with convergence guarantees to a stationary point of the problem. In \cite{Bolte2013}, the authors consider cost functions depending on $N$ blocks of variables of the form:
\begin{equation}
\Psi(\mathbf{x}_1,\ldots,\mathbf{x}_N) := H(\mathbf{x}_1,\ldots,\mathbf{x}_N) + \sum\limits_{j=1}^Nf_j(\mathbf{x}_j) ,
\label{eq:PALMobjective}
\end{equation}
where the function $H$ is smooth, and the $f_j$s are proper and lower semi-continuous (the exact assumptions are given below). It is to be stressed that \emph{no convexity} of any kind is assumed. Here, we assume for simplicity that the $f_j$s are indicator functions of constraint sets $\mathcal{T}_j$. To handle this objective function, the authors propose an algorithm called Proximal Alternating Linearized Minimization (PALM)\cite{Bolte2013}, that updates alternatively each block of variable by a proximal (or projected in our case) gradient step. The structure of the PALM algorithm is given in Algorithm~\ref{algo_summary}, where $P_{\mathcal{T}_j}(\cdot)$ is the projection operator onto the set $\mathcal{T}_j$ and $c^i_j$ defines the step size and depends on the Lipschitz constant of the gradient of $H$ (we give its expression in the next subsection). 
The following conditions are sufficient (not necessary) to ensure that each bounded sequence generated by PALM converges to a stationary point of its objective:
\begin{enumerate}[noitemsep,nolistsep,label=({\roman*})]
\item The $f_j$s are proper and lower semi-continuous.
\item $H$ is smooth.
\item $\Psi$ is semi-algebraic.
\item $\nabla_{\mathbf{x}_j}H$ is globally Lipschitz for all $j$, with Lipschitz moduli $L_j(\mathbf{x}_1\scriptsize{\ldots}\mathbf{x}_{j-1},\mathbf{x}_{j+1}\scriptsize{\ldots}\mathbf{x}_{N})$.
\item $\forall i$, $c^i_j>L_j(\mathbf{x}_1^{i+1}\scriptsize{\ldots}\mathbf{x}_{j-1}^{i+1},\mathbf{x}_{j+1}^{i}\scriptsize{\ldots}\mathbf{x}_{N}^{i})$ (the inequality need not be strict for convex $f_j$).
\end{enumerate}

\begin{algorithm}
\caption{PALM (summary)}
\begin{algorithmic} 
\FOR{$i \in \{1 \cdots Niter\} $} 
\FOR{$j \in \{1 \cdots N\} $}
\STATE Set $\mathbf{x}_j^{i+1} = P_{\mathcal{T}_j}\Big(\mathbf{x}_j^{i} - \frac{1}{c^i_j}\nabla_{\mathbf{x}_j}H\big(\mathbf{x}_1^{i+1}\scriptsize{\ldots}\mathbf{x}_j^{i}\scriptsize{\ldots}\mathbf{x}_{N}^{i}\big)\Big)$
\ENDFOR
\ENDFOR
\end{algorithmic}
\label{algo_summary}
\end{algorithm}
\subsection{Algorithm details}
\label{ssec:algorithmdetails}

 Let us now instantiate PALM for our purpose, namely to handle the objective of eq.\eqref{eq:actualproblemscal}. It is quite straightforward to see that there is a match between eq.\eqref{eq:actualproblemscal} and eq.\eqref{eq:PALMobjective} taking $N = M+2$, $\mathbf{x}_j = \mathbf{S}_j$ for $j \in\{1\ldots M+1\}$, $\mathbf{x}_{M+2} = \lambda$, $H$ is the data fidelity term, $f_j(\cdot) = \delta_{\mathcal{E}_{j}}(.)$ for $j \in\{1\ldots M+1\}$ and $f_{M+2}(\cdot) = \delta_{\mathcal{E}_{M+2}}(\cdot)=\delta_{\mathbb{R}}(\cdot) = 0$ (there is no constraint on $\lambda$). With this particular instance of the problem, conditions (i), (ii) and (iii) are trivially fulfilled provided the $\mathcal{E}_j$s are semi-algebraic sets, which is indeed the case for all the sets considered in this work.

\paragraph{Projection operator.} In the case where the $\mathcal{E}_j$s are defined like in Section~\ref{ssec:objective} with no additional constraints, namely $\mathcal{E}_j = \{\mathbf{A} \in \mathbb{R}^{a_j \times a_{j+1}} : \left\Vert \text{vec}(\mathbf{A}) \right\Vert^0_0 \leq p_j, \left\Vert \mathbf{A} \right\Vert_F = 1\}$ for $j \in\{1\ldots M+1\}$, then the projection operator $P_{\mathcal{E}_j}(\cdot)$ simply keeps the $p_j$ greatest entries (in absolute value) of its argument, sets all the other entries to zero, and then normalize its argument so that it has unit norm (see proof in appendix). Regarding $\mathcal{E}_{M+2} = \mathbb{R}$, the projection operator is the identity mapping. 

\paragraph{Gradient and Lipschitz moduli.} Let us now analyse more precisely the iterations of PALM specialized to our problem. For that we fix the iteration $i$ and the factor $j$. We also need to introduce new notations. First we will call $\mathbf{S}^i := \mathbf{S}^i_j$  the factor that we are updating, $\mathbf{L} := \prod_{k=1}^{j-1} \mathbf{S}^{i+1}_k$ what is on the left of the factor we are updating and $\mathbf{R} := \prod_{k=j+1}^{M+1} \mathbf{S}^{i}_k$ what is on the right (with the convention $\prod_{k\in \varnothing} \mathbf{S}_k = \mathbf{Id}$). Moreover, and to simplify the notation when we update $\lambda$, let us introduce $\hat{\mathbf{X}} = \prod_{k=1}^{M+1} \mathbf{S}^{i+1}_k$. With these new notations we have when updating the $j$th factor:
\(
H(\mathbf{S}_1^{i+1}\ldots\mathbf{S}_j^i\ldots\mathbf{S}_{M+1}^i,\lambda^i) = \tfrac{1}{2}
\| \mathbf{X} - \lambda^i\mathbf{L}\mathbf{S}^i\mathbf{R} 
\|_F^2.
\)
Or equivalently when updating $\lambda$:
\(
H(\mathbf{S}_1^{i+1}\ldots\mathbf{S}_{M+1}^{i+1},\lambda^i) = \tfrac{1}{2}\| \mathbf{X} - \lambda^i\hat{\mathbf{X}} \|_F^2.
\)

 The gradient of this smooth part of the objective with respect to the $j$th factor reads:
\begin{equation*}
\nabla_{\mathbf{S}^i_j}H(\mathbf{S}_1^{i+1}\ldots\mathbf{S}_j^i\ldots\mathbf{S}_{M+1}^i,\lambda^i) = \lambda^i\mathbf{L}^T(\lambda^i\mathbf{L}\mathbf{S}^i\mathbf{R} - \mathbf{X})\mathbf{R}^T,
\end{equation*}
which allows us to verify condition (iv) with $L_j(\mathbf{L},\mathbf{R},\lambda^i) = (\lambda^i)^2\left\Vert \mathbf{R} \right\Vert_2^2. \left\Vert \mathbf{L} \right\Vert_2^2$ (see proof in appendix). 
Fixing a step size $c^i_j$ so as to verify the condition (v), The update of $\mathbf{S}_j^i$ can be rewritten:
\begin{equation*}
\mathbf{S}_j^{i+1} = P_{\mathcal{E}_j}\Big(\mathbf{S}_j^{i} - \frac{1}{c_j^i}\lambda^i\mathbf{L}^T(\lambda^i\mathbf{L}\mathbf{S}_j^i\mathbf{R} - \mathbf{X})\mathbf{R}^T\Big).
\end{equation*}
Now looking at $\lambda$ we have:
\begin{equation*}
\nabla_{\lambda^i}H(\mathbf{S}_1^{i+1}\ldots\mathbf{S}_{M+1}^{i+1},\lambda^i) = \lambda^i \text{Tr}(\hat{\mathbf{X}}^T\hat{\mathbf{X}}) - \text{Tr}(\mathbf{X}^T\hat{\mathbf{X}}). 
\end{equation*}
Since $f_{M+2}(\lambda) = 0$ is a convex penalty, it is enough to check (v) with a non-strict inequality \cite{Bolte2013}, this leads to the update rule:
\begin{equation*}
\lambda^{i+1} =   \frac{\text{Tr}(\mathbf{X}^T\hat{\mathbf{X}})}{\text{Tr}(\hat{\mathbf{X}}^T\hat{\mathbf{X}})}
\end{equation*}

 An explicit version of the algorithm is given in Algorithm~\ref{algo_explicit}. Note that for simplicity we introduce a new notation for the total number of factors $Q:=M+1$.
\begin{algorithm}
\caption{PALM for learning efficient dictionaries (\texttt{palm4LED})}
\begin{algorithmic}[1] 
\REQUIRE{The data matrix $\mathbf{X}$, the desired number of factors $Q$, the constraint sets $\mathcal{E}_j, \: j \in \{1\ldots Q\}$ and a stopping criterion (e.g., here, a number of iterations $N_{iter}$).}
\FOR{$i=0$ to $N_{iter}-1$}
\FOR{$j=1$ to $Q$}
\STATE  $\mathbf{L} \leftarrow \prod_{k=1}^{j-1} \mathbf{S}^{i+1}_k$
\STATE  $\mathbf{R} \leftarrow \prod_{k=j+1}^{Q} \mathbf{S}^{i}_k$
\STATE Set $c^i_j > (\lambda^i)^2\left\Vert \mathbf{R} \right\Vert_2^2. \left\Vert \mathbf{L} \right\Vert_2^2$
\STATE $\mathbf{S}^{i+1}_j \leftarrow P_{\mathcal{E}_j}\Big(\mathbf{S}^{i}_j - \frac{1}{c^i_j}\lambda\mathbf{L}^T(\lambda\mathbf{L}\mathbf{S}^i_j\mathbf{R} - \mathbf{X})\mathbf{R}^T\Big)$
\ENDFOR
\STATE  $\hat{\mathbf{X}} \leftarrow \prod_{k=1}^{Q} \mathbf{S}^{i+1}_k$
\STATE  $\lambda^{i+1} \leftarrow \frac{\text{Tr}(\mathbf{X}^T\hat{\mathbf{X}})}{\text{Tr}(\hat{\mathbf{X}}^T\hat{\mathbf{X}})}$
\ENDFOR
\ENSURE The estimated factorization: $\lambda^{N_{iter}}$,$\{\mathbf{S}^{N_{iter}}_k\}_{k=1}^{Q}$ = \texttt{palm4LED}($\mathbf{X}$, $Q$, $\{\mathcal{E}_j\}_{j=1}^{Q}$)
\end{algorithmic}
\label{algo_explicit}
\end{algorithm}
\vspace{-0.6cm}

\subsection{Practical strategy}
\label{ssec:practical}
Algorithm~\ref{algo_explicit} presented above factorizes a data matrix into sparse factors and converges to a stationary point of the problem stated in eq.\eqref{eq:actualproblemscal}. However, while we are primarily interested in stationary points where the data fidelity term of the cost function is small, there is unfortunately no guarantee that the algorithm converges to such a stationary point.  This is illustrated by a very simple experiment where Algorithm~\ref{algo_explicit} is applied to a data matrix $\mathbf{X} = \mathbf{D}$ with a known factorization in $M$ factors: $\mathbf{D}  = \prod_{j=1}^M \mathbf{S}_j$, such as the Hadamard dictionary.
The naive approach consists in taking directly $Q=M$ in Algorithm~\ref{algo_explicit}, and setting the constraints so as to reflect the actual sparsity of the true factors. This simple strategy performs quite poorly in practice, and the attained local minimum is very often not satisfactory (the data fidelity part of the objective function is big). 




We noticed experimentally that taking fewer factors ($Q$ small) and allowing more non-zero entries per factor led to better results in general. This observation suggested to adopt a hierarchical strategy. Indeed, when $\mathbf{X}  = \prod_{j=1}^{M+1} \mathbf{S}_j$ is the product of $M+1$ sparse factors, it is also the product $\mathbf{X}  = \mathbf{T}_1\mathbf{T}_2$ of $2$ factors $\mathbf{T}_1 = \mathbf{S}_1$ and $\mathbf{T}_2 = \prod_{j=2}^{M+1} \mathbf{S}_j$, so that $\mathbf{T}_1$ is sparser than $\mathbf{T}_2$. Our strategy is then to factorize the data matrix $\mathbf{X}$ in $2$ factors, one being sparse (corresponding to $\mathbf{T}_1$), and the other less sparse (corresponding to $\mathbf{T}_2$). The process can be repeated on the less sparse factor, and so on until we attain the desired number $Q$ of factors. This strategy turns out to be surprisingly effective and the attained local minima are very good, as illustrated in the next section.

 The proposed hierarchical strategy is summarized in Algorithm~\ref{algo_hierarchical}, where we need to specify at each step the constraint sets related to the two factors. For that let us introduce some notation: $\mathcal{E}_k$ will be the constraint set for the left factor and $\tilde{\mathcal{E}}_k$ the one for the right factor at the $k$th factorization. 
The global optimization step (line $5$) is done by initializing \texttt{palm4LED} with the current values of $\{\mathbf{S}_j\}_{j=1}^{k}$ and $\mathbf{R}$. It is here
 to keep an attach to the data matrix $\mathbf{X}$. Roughly we can say that line $3$ of the algorithm is here to yield complexity savings, whereas line $5$ is here to keep low the data fidelity term of the cost function. \\
 {\bf Note:}  the hierarchical strategy can also be applied the other way around (starting {\em from the right}), just by transposing the input. We only present here the version that starts {\em from the left} because the induced notations are simpler.

\begin{algorithm}
\caption{Hierarchical factorization}
\begin{algorithmic}[1]
\REQUIRE{The data matrix $\mathbf{X}$, the desired number of factors $Q$ and the constraint sets $\mathcal{E}_k, \: k \in \{1\ldots Q-1\}$ and $\tilde{\mathcal{E}}_k, \: k \in \{1\ldots Q-1\}$.} 
\STATE $\mathbf{R} \leftarrow \mathbf{X}$
\FOR{$k=1$ to $Q-1$} 
\STATE Factorize the residual $\mathbf{R}$ into $2$ factors: $\lambda'$,$\{\mathbf{T}_1,\mathbf{T}_2\}$ = \texttt{palm4LED}($\mathbf{R}$, $2$, $\{\mathcal{E}_k,\tilde{\mathcal{E}}_k\}$)
\STATE $\mathbf{S}_k \leftarrow \lambda'\mathbf{T}_1$ and $\mathbf{R} \leftarrow \mathbf{T}_2$
\STATE Global optimization: $\lambda$,$\big\{\{\mathbf{S}_j\}_{j=1}^{k},\mathbf{R}\big\}$ = \texttt{palm4LED}($\mathbf{X}$, $k+1$, $\big\{\{\mathcal{E}_j\}_{j=1}^{k},\tilde{\mathcal{E}}_k\big\}$) 
\ENDFOR
\STATE  $\mathbf{S}_Q \leftarrow \mathbf{R}$
\ENSURE The estimated factorization  $\lambda$,$\{\mathbf{S}_k\}_{k=1}^{Q}$.
\end{algorithmic}
\label{algo_hierarchical}
\end{algorithm}
\vspace{-0.2cm}
\section{Experiments}
\label{sec:experiments}
\vspace{-0.3cm}
 In all experiments, we consider square dictionaries and square factors. 
\vspace{-0.3cm}
\subsection{Learning a fast implementation of the Hadamard transform}
\begin{figure}[htbp]
\centering
\includegraphics[width=0.8\textwidth]{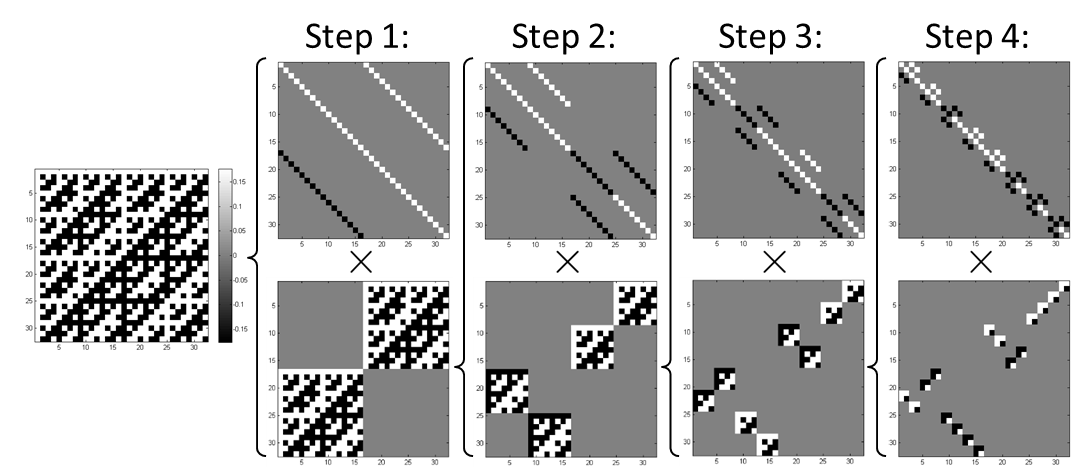}
\caption{Hierarchical factorization of the Hadamard matrix of size $32\times 32$. The matrix is iteratively factorized in 2 factors, until we have $Q=5$ factors, each having $p=64$ non-zero entries.}
\label{fig:hierarchical_strategy}
\end{figure}

\vspace{-0.1cm}
We begin by a dictionary factorization experiment. Consider a data matrix $\mathbf{X} = \mathbf{D}$ with a known factorization in $M$ factors, $\mathbf{D}  = \prod_{j=1}^M \mathbf{S}_j$: in Section~\ref{ssec:practical}, we evoked the failure of Algorithm~\ref{algo_explicit} for this factorization problem. 
In contrast, Figure~\ref{fig:hierarchical_strategy} illustrates the result of the proposed hierarchical strategy (Algorithm~\ref{algo_hierarchical}) with $\mathbf{D}$ the Hadamard dictionary in dimension $n=32$. The obtained factorization is exact and \emph{as good as the reference one} shown on Figure~\ref{fig:hadamard_true} in terms of complexity savings.  The running time is less than a second. Factorization of the Hadamard matrix in dimension up to $n=1024$ showed identical performance, with running time $O(n^{2})$ up to ten minutes.

\vspace{-0.1cm}

\subsection{Learning computationally efficient dictionaries}
\label{ssec:results}
\vspace{-0.1cm}

We now test Algorithm~\ref{algo_hierarchical} in a more realistic framework on a dictionary learning problem with synthetic data, and compare it to classical \emph{learned dictionaries} and \emph{analytic dictionaries}, in terms of approximation quality and relative complexity. 

\renewcommand{\arraystretch}{0.5} 
\vspace{-0.1cm}
\paragraph{Data.} To build the data matrix $\mathbf{X}$, we generated $500$ training samples by selecting uniformly at random $5$ atoms in a dictionary $\mathbf{D}_0 \in \mathbb{R}^{32 \times 32}$ with i.i.d. Gaussian coefficients to build each sample. Gathering the coefficients in the matrix $\boldsymbol{\Gamma}_0 \in \mathbb{R}^{32 \times 500}$ we get our training data matrix $\mathbf{X} = \mathbf{D}_0\boldsymbol{\Gamma}_0 \in \mathbb{R}^{32 \times 500}$. Two reference dictionaries $\mathbf{D}_0$ are considered:
\begin{itemize}[noitemsep,nolistsep,leftmargin=0.3cm]
\item {\em Factorizable dictionary} (FACT):  $\mathbf{D}_0$ is the product of $M=5$ sparse matrices: $\mathbf{D}_0 = \prod_{i=1}^5 \mathbf{S}_i^0$, each having a random number $p_i^0$ of i.i.d. Gaussian non-zero entries with $64\leq p_i^0 \leq 128$, and being full rank so that the dictionary spans the signal space.
\item {\em Random dictionary} (RAND): $\mathbf{D}_0$ has i.i.d. Gaussian entries.
\end{itemize} 
\vspace{-0.1cm}
\paragraph{Baselines.} We compare Algorithm~\ref{algo_hierarchical} with the following methods. All methods involve a coefficient update step which is performed using Orthogonal Matching Pursuit (OMP) \cite{Mallat1993}: 
\begin{itemize}[noitemsep,nolistsep,leftmargin=0.3cm]
\item K-SVD \cite{Aharon2006}, one of the most used algorithm that provides a \emph{learned dictionary}. We use the implementation described in \cite{Rubinstein2008}, running $300$ iterations (which proved empirically sufficient to ensure convergence). Note that we also tested the online dictionary learning (ODL) method of \cite{Mairal2010}. Its performance being almost identical to that of K-SVD in our setting, we decided to consider these two methods as one (K-SVD/ODL) in the interpretation of the results.
\item Sparse K-SVD \cite{Rubinstein2010a}, a method that seeks to bridge the gap between \emph{learned dictionaries} and \emph{analytic dictionaries}. The implementation of \cite{Rubinstein2010a} is used, with $\mathbf{D}_{\textrm{base}}$ the Discrete Cosine Transform (DCT) matrix and  $100$ iterations, ensuring convergence in practice. The estimated dictionary $\mathbf{D}$ has columns $4$-sparse in $\mathbf{D}_{\textrm{base}}$. 
\item A fixed \emph{analytic dictionary} with a known fast implementation (either the DCT, the Haar wavelets (HAAR) or the Hadamard matrix (HAD)).
\end{itemize}
\vspace{-0.1cm}
\paragraph{Settings of our algorithm.}
We tested several configurations for Algorithm~\ref{algo_hierarchical}, and we present here only the best one (PROPOSED). It amounts to Algorithm~\ref{algo_hierarchical} starting from the right, with two modifications. First, we performed the first factorization ($k=1$) by  K-SVD/ODL to compute $\mathbf{S}_{M+1} = \boldsymbol{\Gamma}$. Second,  we noticed that it was beneficial to update the coefficient matrix $\boldsymbol{\Gamma}$ with OMP after each global optimization step of Algorithm~\ref{algo_hierarchical} (line 5). 
We tested various numbers of factors $Q \in \{3\ldots 6\}$. The considered constraint sets were:  $\mathcal{E}_1 = \{\mathbf{A} \in \mathbb{R}^{32 \times 500}, \left\Vert \mathbf{a}_n \right\Vert_0^0 \leq 5\}$ and for $k \in \{2\ldots Q-1\}$, $\mathcal{E}_k = \{\mathbf{A} \in \mathbb{R}^{32 \times 32}, \left\Vert \text{vec}(\mathbf{A}) \right\Vert_0^0 \leq p, \left\Vert \mathbf{A} \right\Vert_F = 1 \}$ and $\tilde{\mathcal{E}}_k = \{\mathbf{A} \in \mathbb{R}^{32 \times 32}, \left\Vert \text{vec}(\mathbf{A}) \right\Vert_0^0 \leq \frac{P}{2^{k-2}}, \left\Vert \mathbf{A} \right\Vert_F = 1 \}$. We show the results for sparsity constraints given by $p \in \{2,3,4\}$ and $P \in 512\times\{1,1.2,1.4,1.6\}$. Algorithm~\ref{algo_hierarchical} was implemented  in Matlab, and executed on Intel Core i7-3667U CPU. It typically took between 8 and 9 seconds to converge for each drawn $\mathbf{X}$.
The stopping criterion for $\texttt{palm4LED}$ combined a maximum number of iterations $N_{tier}=500$ and a bound $\epsilon=10^{-6}$ on the variation of the approximation error between consecutive iterations.
\vspace{-0.1cm}
\paragraph{Performance measures.}
\label{ssec:perfs}
The ideal dictionary should approximate well the data at hand, while being at the same time fast to manipulate and cheap to store. 
The computational efficiency of the dictionary is measured through the \emph{Relative complexity} (RC) quantity introduced in Section~\ref{ssec:objective}.
The quality of approximation is expressed using the Root-Mean-Square Error (RMSE)\cite{Rubinstein2010a, Aharon2006}:
\(
{\text{RMSE} := \tfrac{1}{\sqrt{dn}}\left\Vert \mathbf{X} - \mathbf{D}\boldsymbol{\Gamma} \right\Vert_F}.
\)

\begin{figure}[htbp]
\centering
\includegraphics[width=1\textwidth]{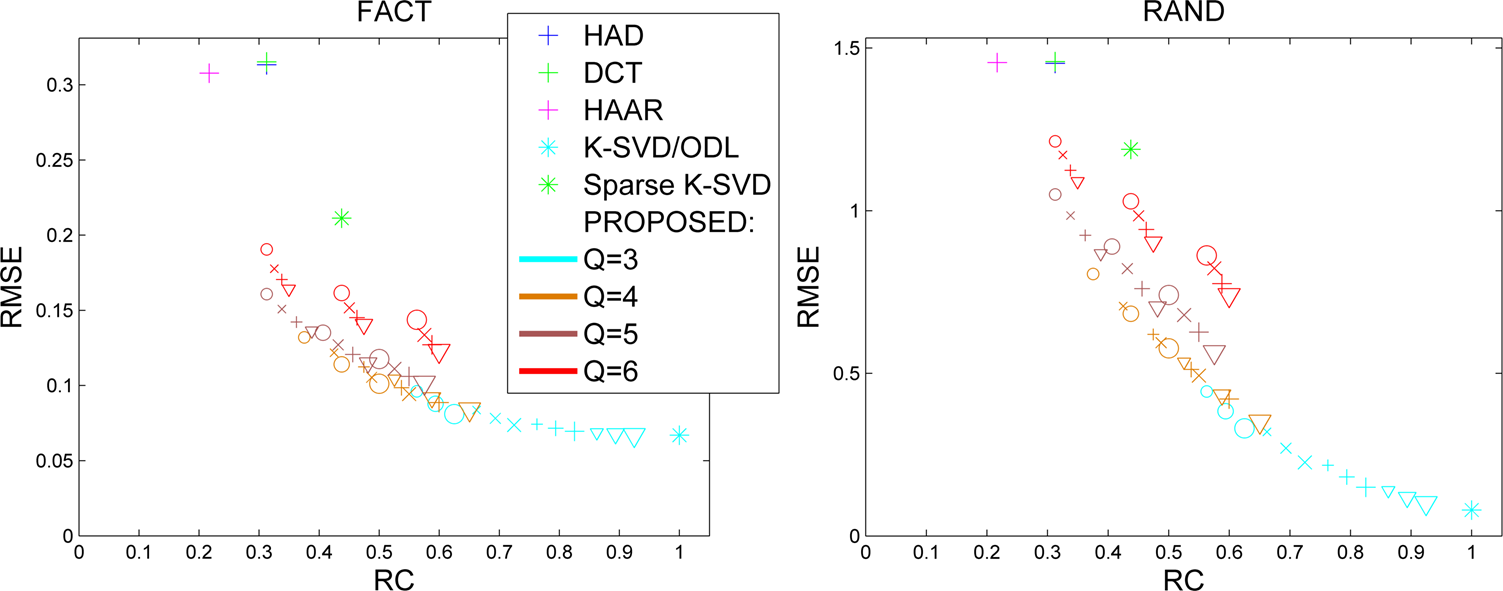}
\caption{(better seen in colors) Comparison of different dictionary learning methods, with data generated using different dictionaries:  factorizable (left), random (right).
For the proposed method, the symbol shape indicates the value of $P$ ($\fullmoon$ : $P=512$, $\times$: $P=1.2\times 512$, $+$: $P=1.4\times 512$ and $\triangledown$: $P=1.6\times 512$), its color the value of $Q$ (see legend), and its size the value of $p$ (Small: $p=2$, Medium: $p=3$, Big: $p=4$).}
\label{fig:results}
\end{figure}


\vspace{-0.28cm}\paragraph{Discussion of the results.} The experiment has been repeated $100$ times with each data generation method. For a given configuration of the algorithm the relative complexity is constant over all trials, and the results shown on Figure~\ref{fig:results} display the average RMSE. 

With a {\em factorizable dictionary} (left), as expected, the methods that use a fast dictionary (HAD, DCT and HAAR) perform quite poorly in approximation (vertical axis), but very good in relative complexity (horizontal axis) taking advantage of their intrinsic structure. On the other hand, K-SVD exhibits good approximation performance, while the lack of structure of the obtained dictionary does not lead to any complexity savings ($\text{RC}=1$). In between these two extremes, Sparse K-SVD, thanks to its layer of adaptivity, performs better than the analytic dictionaries in approximation at the expense of a slightly higher relative complexity. The proposed method (PROPOSED) has the ability to achieve a flexible tradeoff between complexity and adaptation to the data. More specifically, we can identify several behaviors for the proposed method. With $Q=3$ factors and $P = 1.4\times512$ or $P=1.6\times512$, the proposed method performs almost as good as K-SVD in terms of approximation, with reduced relative complexities between $0.7$ and $0.9$. On the other hand, with $p = 2$ and $Q=5$ or $Q=6$, the proposed method provides dictionaries almost as compact as analytic dictionaries ($\text{RC}\approx0.3$), while being better adapted to the data (RMSE up to twice smaller). The other configurations of $p$, $P$ and $Q$ all lie between these two behaviors in terms of performance.

With a {\em random dictionary} (right), the methods exhibit qualitatively the same comparative behavior as with a factorizable dictionary. 
Notably, the proposed method can learn a dictionary as computationally efficient as the one provided by Sparse K-SVD but with half the approximation error.


\vspace{-0.2cm}
\section{Conclusion}
\label{sec:conclusion}
\vspace{-0.3cm}
We proposed a dictionary learning framework that provides a flexible tradeoff between computational efficiency and adaptation to the training data. Stemming on recent advances in non-convex optimization, we derived an algorithm with convergence guarantees to learn efficient dictionaries and demonstrated experimentally its ability to provide complexity/accuracy tradeoffs that state of the art dictionary learning methods could not achieve. Besides the obvious need to further test the approach on real data and with redundant dictionaries, and to better understand the role of its parameters in the control of the desired tradeoff, a particular challenge will be to leverage the gained complexity to speed up the learning process itself, in order to {\em efficiently learn} efficient dictionaries. 

\vspace{-0.1cm}
\subsubsection*{Acknowledgments}
\vspace{-0.1cm}
This work was supported in part by the European Research Council, PLEASE project (ERC-StG- 2011-277906).
{ The authors wish to thank Fran\c{c}ois Malgouyres and Olivier Chabiron for the fruitful discussions that helped in producing that work. }
\newpage
\subsubsection*{References}
\begingroup
\renewcommand{\section}[2]{}%
\bibliographystyle{unsrt}
\bibliography{biblio}
\endgroup

\appendix
\section{Factorizations examples}
In this appendix we show that the matrices of usual transforms associated with fast algorithm can be factorized into sparse factors.
\subsection{The Discrete Fourier Transform}  
We are going to look at the DFT matrix in dimension $8$ and show that it can be factorized into sparse matrices. Note that a similar factorization can be done in any power of two dimension. Let us take $\mathbf{C} \in \mathbb{C}^{8 \times 8}$ to be the DFT matrix:
\begin{equation}\def\arraystretch{1.5}
   \mathbf{C} = \left( \begin{array}{cccccccc}
      W^0 & W^0 & W^0 & W^0 & W^0 & W^0 & W^0 & W^0   \\
      W^0 & W^1 & W^2 & W^3 & W^4 & W^5 & W^6 & W^7   \\
      W^0 & W^2 & W^4 & W^6 & W^0 & W^2 & W^4 & W^6   \\
      W^0 & W^3 & W^6 & W^1 & W^4 & W^7 & W^2 & W^5   \\
      W^0 & W^4 & W^0 & W^4 & W^0 & W^4 & W^0 & W^4   \\
      W^0 & W^5 & W^2 & W^7 & W^4 & W^1 & W^6 & W^3   \\
      W^0 & W^6 & W^4 & W^2 & W^0 & W^6 & W^4 & W^2   \\
      W^0 & W^7 & W^6 & W^5 & W^4 & W^3 & W^2 & W^1   \\
    \end{array}
    \right),
\end{equation}    
with $W = \exp(\frac{2\pi i}{8})$. Applying permutations of rows and columns (bit-reversed order), we obtain: 
\begin{equation}\def\arraystretch{1.5}
   \mathbf{P}_r\mathbf{C}\mathbf{P}_c = \left( \begin{array}{cccccccc}
      W^0 & W^0 & W^0 & W^0 & W^0 & W^0 & W^0 & W^0   \\
      W^0 & W^0 & W^0 & W^0 & W^4 & W^4 & W^4 & W^4   \\
      W^0 & W^0 & W^4 & W^4 & W^2 & W^2 & W^6 & W^6   \\
      W^0 & W^0 & W^4 & W^4 & W^6 & W^6 & W^2 & W^2   \\
      W^0 & W^4 & W^2 & W^6 & W^1 & W^5 & W^3 & W^7   \\
      W^0 & W^4 & W^2 & W^6 & W^5 & W^1 & W^7 & W^3   \\
      W^0 & W^4 & W^6 & W^2 & W^3 & W^7 & W^1 & W^5   \\
      W^0 & W^4 & W^6 & W^2 & W^7 & W^3 & W^5 & W^1   \\
    \end{array}
    \right).
\end{equation}

This matrix can be factorized as follows:
\begin{equation}\def\arraystretch{1.5}
\small{
   \mathbf{P}_r\mathbf{C}\mathbf{P}_c = \left( \begin{array}{cccccccc}
      W^0 & W^0 &  0  &  0  &  0  &  0  &  0  &  0    \\
      W^0 & W^4 &  0  &  0  &  0  &  0  &  0  &  0    \\
       0  &  0  & W^0 & W^0 &  0  &  0  &  0  &  0    \\
       0  &  0  & W^0 & W^4 &  0  &  0  &  0  &  0    \\
       0  &  0  &  0  &  0  & W^0 & W^0 &  0  &  0    \\
       0  &  0  &  0  &  0  & W^0 & W^4 &  0  &  0    \\
       0  &  0  &  0  &  0  &  0  &  0  & W^0 & W^0   \\
       0  &  0  &  0  &  0  &  0  &  0  & W^0 & W^4   \\
    \end{array}
    \right) \times
    \left( \begin{array}{cccccccc}
      W^0 & W^0 & W^0 & W^0 &  0  &  0  &  0  &  0    \\
       0  &  0  &  0  &  0  & W^0 & W^0 & W^0 & W^0   \\
      W^0 & W^0 & W^4 & W^4 &  0  &  0  &  0  &  0    \\
       0  &  0  &  0  &  0  & W^2 & W^2 & W^6 & W^6   \\
      W^0 & W^4 & W^2 & W^6 &  0  &  0  &  0  &  0    \\
       0  &  0  &  0  &  0  & W^1 & W^5 & W^3 & W^7   \\
      W^0 & W^4 & W^6 & W^2 &  0  &  0  &  0  &  0    \\
       0  &  0  &  0  &  0  & W^3 & W^7 & W^5 & W^1   \\
    \end{array}
    \right)}.
\end{equation}
At this point the left factor can be further factorized:
\begin{equation}\def\arraystretch{1.5}
\begin{array}{c}

   \mathbf{P}_r\mathbf{C}\mathbf{P}_c = \left( \begin{array}{cccccccc}
      W^0 & W^0 &  0  &  0  &  0  &  0  &  0  &  0    \\
      W^0 & W^4 &  0  &  0  &  0  &  0  &  0  &  0    \\
       0  &  0  & W^0 & W^0 &  0  &  0  &  0  &  0    \\
       0  &  0  & W^0 & W^4 &  0  &  0  &  0  &  0    \\
       0  &  0  &  0  &  0  & W^0 & W^0 &  0  &  0    \\
       0  &  0  &  0  &  0  & W^0 & W^4 &  0  &  0    \\
       0  &  0  &  0  &  0  &  0  &  0  & W^0 & W^0   \\
       0  &  0  &  0  &  0  &  0  &  0  & W^0 & W^4   \\
    \end{array}
    \right) 
    \\
    \\
    \times
    \left( \begin{array}{cccccccc}
      W^0 & W^0 &  0  &  0  &  0  &  0  &  0  &  0    \\
       0  &  0  & W^0 & W^0 &  0  &  0  &  0  &  0    \\
      W^0 & W^4 &  0  &  0  &  0  &  0  &  0  &  0    \\
       0  &  0  & W^2 & W^6 &  0  &  0  &  0  &  0    \\
       0  &  0  &  0  &  0  & W^0 & W^0 &  0  &  0    \\
       0  &  0  &  0  &  0  &  0  &  0  & W^0 & W^0   \\
       0  &  0  &  0  &  0  & W^0 & W^4 &  0  &  0    \\
       0  &  0  &  0  &  0  &  0  &  0  & W^2 & W^6   \\
    \end{array}
    \right)
\\ 
\\   
    \times
    \left( \begin{array}{cccccccc}
      W^0 & W^0 &  0  &  0  &  0  &  0  &  0  &  0    \\
       0  &  0  & W^0 & W^0 &  0  &  0  &  0  &  0    \\
       0  &  0  &  0  &  0  & W^0 & W^0 &  0  &  0    \\
       0  &  0  &  0  &  0  &  0  &  0  & W^0 & W^0   \\
      W^0 & W^4 &  0  &  0  &  0  &  0  &  0  &  0    \\
       0  &  0  & W^2 & W^6 &  0  &  0  &  0  &  0    \\
       0  &  0  &  0  &  0  & W^1 & W^5 &  0  &  0    \\
       0  &  0  &  0  &  0  &  0  &  0  & W^3 & W^7   \\
    \end{array}
    \right)
    \end{array},
\end{equation}
This factorization actually corresponds to the butterfly radix-2 FFT.

\subsection{The Hadamard transform}
\begin{figure}[h!]
\includegraphics[width=1\textwidth]{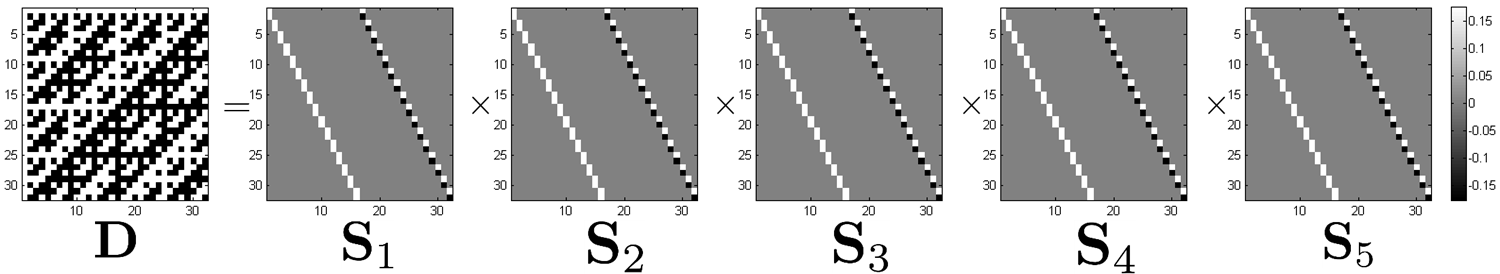}
\caption{The Hadamard dictionary in size $n\times n$ with $n=32$ (left) and its factorization. The dictionary is totally dense so that the naive storage and multiplication cost $\mathcal{O}(n^2=1024)$. On the other hand, we show the factorization of the dictionary in $\log_2(n) = 5$ factors, each having $2n = 64$ non-zero entries, so that the storage and multiplication in the factorized form cost \mbox{$\mathcal{O}(2n\log_2(n)=320)$.}}
\label{fig:hadamard_true}
\end{figure}

\subsection{The wavelet transform}
We are interested in the factorization's structure of the Discrete Wavelet Transform (DWT) matrix. More precisely, we wish to express the synthesis of a signal $\mathbf{x} \in  \mathbb{R}^{2^M}$ from its wavelet coefficients $\boldsymbol\gamma \in  \mathbb{R}^{2^M}$ as a sequence of simple linear transformations, i.e.\ a product of $\boldsymbol\gamma$ by multiple sparse matrices.

The discrete signal $\mathbf{x}$ can be seen as the coordinates $\mathbf{a}_M$ in a basis $\{\phi_{j,M}\}_{j=1}^{2^M}$ of the projection of the underlying continuous signal onto the approximation space $V_M$. Multi Resolution Analysis (MRA) consists in defining a hierarchy of subspaces : $V_M \supset V_{M-1} \supset \cdots \supset V_{0}$ and their direct complement $V_k = V_{k-1} \oplus W_{k-1}$ such that $V_{k-1}$ and $W_{k-1}$ are of dimension $2^{k-1}$. By induction we have: $V_M = V_{0} \oplus W_{0} \oplus \cdots \oplus W_{M-1}$. The DWT is then a change of basis from the canonical basis to a basis which is the union of bases from each subspace $ V_{0} , W_{0} , \cdots , W_{M-1} $.


We define $\boldsymbol\gamma = (\mathbf{a}_0^T|\mathbf{b}_0^T|\mathbf{b}_1^T|\cdots|\mathbf{b}_{M-1}^T)^T$ , where $\mathbf{a}_k \in  \mathbb{R}^{2^k}$, $\mathbf{b}_k \in  \mathbb{R}^{2^k}$ as the DWT of a signal $\mathbf{x}=\mathbf{a}_M \in  \mathbb{R}^{2^M}$ that can be obtained by $M$ iterations of the following filterbank: 

 \begin{figure}[htbp]
\center
    \begin{tikzpicture}[>=latex']
        \tikzset{block/.style= {draw, rectangle, align=center,minimum width=1cm,minimum height=1cm},
        rblock/.style={draw, shape=rectangle,rounded corners=1.5em,align=center,minimum width=2cm,minimum height=1cm},
        cblock/.style= {draw, circle, align=center,minimum width=0.5cm,minimum height=0.5cm},
        input/.style={ 
        draw,
        trapezium,
        trapezium left angle=60,
        trapezium right angle=120,
        minimum width=2cm,
        align=center,
        minimum height=1cm
    },
        }
        \node [circle]  (start) {$\mathbf{a}_{k}$};
        \node [circle, below =1cm of start]  (start2) {$\mathbf{b}_{k}$};
        \node [block, left =1cm of start] (upsample1) { $\downarrow2$};
        \node [block, left =1cm of start2] (upsample2) { $\downarrow2$};
        \node [block, left =0.7cm of upsample1] (filterg) {$e$};
        \node [block, left =0.7cm of upsample2] (filterh) {$f$};
        \node [circle, below left =0.0cm and 0.7cm of filterg]  (plus) {$\mathbf{a}_{k+1}$};

        \path[draw,->] (upsample1) edge (start)
                    (upsample2) edge (start2)
                    (plus) edge (filterg)
                    (plus) edge (filterh)
                    (filterg) edge (upsample1)
                     (filterh)edge (upsample2)

                    ;
    \end{tikzpicture}
    \caption{Signal analysis elementary block}
    \end{figure}
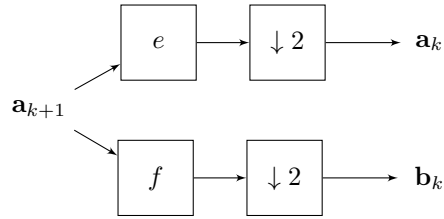

The inverse transform is obtained by $M$ iterations of this other filterbank: 

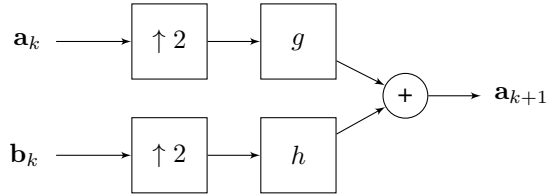
\begin{figure}[htbp]
\center
    \begin{tikzpicture}[>=latex']
        \tikzset{block/.style= {draw, rectangle, align=center,minimum width=1cm,minimum height=1cm},
        rblock/.style={draw, shape=rectangle,rounded corners=1.5em,align=center,minimum width=2cm,minimum height=1cm},
        cblock/.style= {draw, circle, align=center,minimum width=0.5cm,minimum height=0.5cm},
        input/.style={ 
        draw,
        trapezium,
        trapezium left angle=60,
        trapezium right angle=120,
        minimum width=2cm,
        align=center,
        minimum height=1cm
    },
        }
        \node [circle]  (start) {$\mathbf{a}_{k}$};
        \node [block, right =1cm of start] (upsample1) { $\uparrow2$};
        \node [block, below =0.5cm of upsample1] (upsample2) { $\uparrow2$};
        \node [block, right =0.7cm of upsample1] (filterg) {$g$};
        \node [block, right =0.7cm of upsample2] (filterh) {$h$};
                \node [circle, left =1cm of upsample2]  (start2) {$\mathbf{b}_{k}$};
        \node [cblock, below right =0.0cm and 0.7cm of filterg]  (plus) {+};
        \node [circle, right =0.7cm of plus]  (end) {$\mathbf{a}_{k+1}$};

        \path[draw,->] (start) edge (upsample1)
                    (start2) edge (upsample2)
                    (filterg) edge (plus)
                    (filterh) edge (plus)
                    (upsample1) edge (filterg)
                    (upsample2) edge (filterh)
                    (plus) edge (end)
                    ;
    \end{tikzpicture}
    \label{fig:synthesis}
    \caption{Signal synthesis elementary block}
    \end{figure}

Downsampling, upsampling and filtering being linear transformations, the synthesis and analysis elementary blocks can be seen as matrix products. Let us focus on the synthesis case. We define $\mathbf{M}_k \in \mathbb{R}^{2^{k} \times 2^{k}} $ such that $\mathbf{a}_{k+1} = \mathbf{M}_{k+1} .(\mathbf{a}_{k}^T|\mathbf{b}_{k}^T)^T$. The matrix $\mathbf{M}_k$ accounts for upsampling followed by filtering with two different filters, so it takes the following form: 
\begin{equation}
\mathbf{M}_k = \left(\begin{array}{c|c}
\mathbf{G}_k & \mathbf{H}_k
\end{array}\right)
. \left(\begin{array}{cc}
\mathbf{U}_k & \mathbf{0} \\ \mathbf{0} & \mathbf{U}_k
\end{array}\right) = \left(\begin{array}{c|c}
\mathbf{G}_{k\downarrow2} & \mathbf{H}_{k\downarrow2}
\end{array}\right),
\end{equation}
it is the concatenation of two columnwise downsampled Toeplitz (or circulant) matrices.

Introducing $\mathbf{Id}_k \in \mathbb{R}^{(2^M - 2^k)\times (2^M - 2^k)}$ the identity in dimension $2^M - 2^k$, and the matrix $\mathbf{S}_k \in \mathbb{R}^{2^M\times2^M}$ taking the form: 
\begin{equation}
\mathbf{S}_k = \left(\begin{array}{cc}
\mathbf{M}_k&\mathbf{0}\\ 
\mathbf{0}&\mathbf{Id}_k
\end{array}\right),
\label{eq:structure}
\end{equation}
the inverse DWT of $\boldsymbol\gamma$ can be expressed:

\begin{equation}
\mathbf{x} = \prod_{k=1}^M \mathbf{S}_k \boldsymbol\gamma
\end{equation}

\end{document}


\maketitle

\section{Projection operator proof}
We want to find the projection operator onto the following set:
\begin{equation}
\mathcal{E} := \{\mathbf{A} \in \mathbb{R}^{n \times n} : \left\Vert \mathbf{A} \right\Vert_0  \leq p, \left\Vert \mathbf{A} \right\Vert_F  = 1 \},
\end{equation}
with $p \in \mathbb{N}^*$.
We are interested in the projection of some matrix $\mathbf{S} \in \mathbb{R}^{n \times n}$ onto the set $\mathcal{E}$, namely we want to find $\mathbf{S}^*$ such that:
\begin{equation}
\mathbf{S}^* = P_{\mathcal{E}}(\mathbf{S})\in \underset{\mathbf{U}}{\arg\min}\{ \left\Vert \mathbf{U} - \mathbf{S} \right\Vert_F^2 : \mathbf{U} \in \mathcal{E}\}
\end{equation}

\begin{proposition}
Projection operator formula.

\begin{equation*}
P_{\mathcal{E}}(\mathbf{S}) = B_1(T_p(\mathbf{S}))
\end{equation*}
where 
\begin{equation*}
T_p(\mathbf{S}) := \underset{\mathbf{V}}{\arg\min}\{ \left\Vert \mathbf{V} - \mathbf{S} \right\Vert_F^2 : \left\Vert \mathbf{V} \right\Vert_0 \leq p \}
\end{equation*} 
and 
\begin{equation*}
B_1(\mathbf{R}) := \underset{\mathbf{W}}{\arg\min}\{ \left\Vert \mathbf{W} - \mathbf{R} \right\Vert_F^2 : \left\Vert \mathbf{W} \right\Vert_F = 1 \}
\end{equation*}
\end{proposition}

\begin{proof}
Let us denote by $P$ the set of indices corresponding to the $p$ greatest entries (in absolute value) of $\mathbf{S}$, and by $\mathbf{S}_P$ the matrix we get by setting all the other entries of $\mathbf{S}$ to $0$. Let us also introduce $\mathbf{S}_{\bar{P}} = \mathbf{S}- \mathbf{S}_P$, the matrix we get by keeping only the $n^2-p$ littlest entries.
We have:
\begin{equation}
\left\Vert \mathbf{S} \right\Vert_F^2 = \left\Vert \mathbf{S}_P \right\Vert_F^2 + \left\Vert \mathbf{S}_{\bar{P}} \right\Vert_F^2
\end{equation} 
and 
\begin{equation}
\mathbf{S}_P  = \underset{\mathbf{V}}{\arg\min}\{ \left\Vert \mathbf{V} - \mathbf{S} \right\Vert_F^2 : \left\Vert \mathbf{V} \right\Vert_0 \leq p \} = T_p(\mathbf{S})
\end{equation} 
and 
\begin{equation}
P  = \underset{J}{\arg\max}\{ \left\Vert \mathbf{S}_J \right\Vert_F : \text{card}(J) \leq p \} 
\end{equation}
The operator $B_1$ is simply the projection onto the Euclidean sphere of radius $1$, that is:
\begin{equation}
B_1(\mathbf{X}) = \frac{1}{\left\Vert \mathbf{X}\right\Vert_F} \mathbf{X}.
\end{equation}

Let us now denote by $I$ the support of any $\mathbf{U} \in \mathcal{E}$, we have:
\begin{equation}
\left\Vert \mathbf{U} - \mathbf{S} \right\Vert_F^2 = \left\Vert \mathbf{S}_{\bar{I}} \right\Vert_F^2 + \left\Vert \mathbf{U} - \mathbf{S}_I \right\Vert_F^2
\end{equation}
so the problem can be rewritten: 
\begin{equation}
\begin{array}{rl}
(\mathbf{S}^*,I^*) = &
\underset{(\mathbf{U},I)}{\arg\min}\{\left\Vert \mathbf{S}_{\bar{I}} \right\Vert_F^2 +   \left\Vert \mathbf{U} - \mathbf{S}_I \right\Vert_F^2 : \left\Vert \mathbf{U}\right\Vert_F = 1 ,  \text{card}(I) \leq p \}
\end{array}
\end{equation}
Isolating $\mathbf{U}$ we have:
\begin{equation}
\mathbf{S}^* =\underset{\mathbf{U}}{\arg\min}\{ \left\Vert \mathbf{U} - \mathbf{S}_{I^*} \right\Vert_F^2 : \left\Vert \mathbf{U} \right\Vert_F \leq q \} =B_1(\mathbf{S}_{I^*})
\end{equation}
with:
\begin{equation}
\begin{array}{rl} 
I^* = &
\underset{I}{\arg\min}\{\left\Vert \mathbf{S}_{\bar{I}} \right\Vert_F^2 + \underset{\mathbf{U}}{\text{inf}}\{  \left\Vert \mathbf{U} - \mathbf{S}_I \right\Vert_F^2 : \left\Vert \mathbf{U}\right\Vert_F = 1 \} :  \text{card}(I) \leq p \}
\\
I^*=&\underset{I}{\arg\min}\{\left\Vert \mathbf{S}_{\bar{I}} \right\Vert_F^2 + \left\Vert B_1(\mathbf{S}_I) - \mathbf{S}_I\right\Vert_F^2 :  \text{card}(I) \leq p \} 
\\
I^*=&\underset{I}{\arg\min}\{\left\Vert \mathbf{S}_{\bar{I}} \right\Vert_F^2 +(1-\frac{1}{\left\Vert \mathbf{S}_I\right\Vert_F})^2 \left\Vert \mathbf{S}_I\right\Vert_F^2 :  \text{card}(I) \leq p \} 
\\
I^*=&\underset{I}{\arg\min}\{\underbrace{\left\Vert \mathbf{S}_{\bar{I}} \right\Vert_F^2 + \left\Vert \mathbf{S}_{I} \right\Vert_F^2}_{\left\Vert \mathbf{S} \right\Vert_F^2} -2 \left\Vert \mathbf{S}_I\right\Vert_F :  \text{card}(I) \leq p \} 
\\
I^*=&\underset{I}{\arg\max}\{ \left\Vert \mathbf{S}_I\right\Vert_F :  \text{card}(I) \leq p \} 
\\
I^*=&P
\end{array}
\end{equation}
So we have: $\mathbf{S}^* = B_1(\mathbf{S}_P) = B_1(T_p(\mathbf{S}))$.
\end{proof}

\section{Lipschitz moduli proof}
Let us look at the Lipschitz moduli of the gradient of the smooth part of the objective:
\begin{equation}
\begin{array}{ll}
&\left\Vert \nabla_{\mathbf{S^i_j}}H(\mathbf{S}_1^{i+1}\ldots\mathbf{S}_1\ldots\mathbf{S}_p^i,\lambda^i) - \nabla_{\mathbf{S^i_j}}H(\mathbf{S}_1^{i+1}\ldots\mathbf{S}_2\ldots\mathbf{S}_p^i,\lambda^i)\right\Vert_F \\ \\
=& \left\Vert\lambda^i\mathbf{L}^T(\lambda^i\mathbf{L}\mathbf{S}_1\mathbf{R} - \mathbf{X})\mathbf{R}^T - \lambda^i\mathbf{L}^T(\lambda^i\mathbf{L}\mathbf{S}_2\mathbf{R} - \mathbf{X})\mathbf{R}^T\right\Vert_F \\ \\
=& (\lambda^i)^2 \left\Vert \mathbf{L}^T\mathbf{L} (\mathbf{S}_1-\mathbf{S}_2) \mathbf{R}\mathbf{R}^T\right\Vert_F\\ \\
=& (\lambda^i)^2 \left\Vert (\mathbf{R}\mathbf{R}^T)\otimes(\mathbf{L}^T\mathbf{L}) . \text{vec}(\mathbf{S}_1-\mathbf{S}_2)\right\Vert_2\\ \\
\leq& (\lambda^i)^2 \left\Vert (\mathbf{R}\mathbf{R}^T)\otimes(\mathbf{L}^T\mathbf{L})\right\Vert_2 \left\Vert\mathbf{S}_1-\mathbf{S}_2 \right\Vert_F\\ \\
=& (\lambda^i)^2\left\Vert \mathbf{R} \right\Vert_2^2. \left\Vert \mathbf{L} \right\Vert_2^2 \left\Vert\mathbf{S}_1-\mathbf{S}_2 \right\Vert_F.
\end{array}
\end{equation}
So we can say that the following quantity is a Lipschitz modulus:
$L_j(\mathbf{L},\mathbf{R},\lambda^i) = (\lambda^i)^2\left\Vert \mathbf{R} \right\Vert_2^2. \left\Vert \mathbf{L} \right\Vert_2^2$.

\bibliographystyle{plain}
\bibliography{bibli}